\documentclass[sigconf,nonacm]{acmart}

\usepackage{amsmath,array,booktabs,graphicx,microtype,placeins,xspace}
\newtheorem{assumption}{Assumption}
\newtheorem{proposition}{Proposition}
\newtheorem{theorem}{Theorem}
\newcommand{\F}{\mathcal F}
\newcommand{\E}{\mathbb E}
\newcommand{\Pp}{\mathbb P}
\newcommand{\PROMOTE}{\textsc{Promote}\xspace}

\newcommand{\RuleCount}{24}
\newcommand{\NullReps}{3,000}
\newcommand{\NaiveFWER}{94.5\%}
\newcommand{\SafeFWER}{0.0\%}
\newcommand{\SignalPower}{70.1\%}
\newcommand{\MedianDelay}{2,327}
\newcommand{\RealPromotions}{zero}
\newcommand{\AdaptiveReps}{3,000}
\newcommand{\AdaptiveBirths}{9.88}
\newcommand{\AdaptiveFWER}{1.77\%}

\setcopyright{none}
\makeatletter\global\@ACM@balancefalse\makeatother
\acmConference{}{}{}
\acmBooktitle{}
\acmDOI{}
\acmISBN{}

\begin{document}

\title[Safe Promotion Tests for Financial Research Agents]{Can We Stop the Backtest Yet? Anytime-Valid Promotion Tests for Adaptive Financial Research Agents}

\author{Eric Benhamou}
\affiliation{
  \institution{Ai For Alpha}
  \city{Neuilly-sur-Seine}
  \country{France}
}
\affiliation{
  \institution{LAMSADE, Universit\'e Paris Dauphine--PSL}
  \city{Paris}
  \country{France}
}
\email{eric.benhamou@aiforalpha.com}

\author{Jean-Jacques Ohana}
\affiliation{
  \institution{Ai For Alpha}
  \city{Neuilly-sur-Seine}
  \country{France}
}
\email{jean-jacques.ohana@aiforalpha.com}

\author{Beatrice Guez}
\affiliation{
  \institution{Ai For Alpha}
  \city{Neuilly-sur-Seine}
  \country{France}
}
\email{beatrice.guez@aiforalpha.com}

\author{Ethan Setrouk}
\affiliation{
  \institution{Ai For Alpha}
  \city{Neuilly-sur-Seine}
  \country{France}
}
\email{ethan.setrouk@aiforalpha.com}

\author{Chamyl Saadi}
\affiliation{
  \institution{Ai For Alpha}
  \city{Neuilly-sur-Seine}
  \country{France}
}
\email{chamyl.saadi@aiforalpha.com}

\author{Thomas Jacquot}
\affiliation{
  \institution{Ai For Alpha}
  \city{Neuilly-sur-Seine}
  \country{France}
}
\email{thomas.jacquot@aiforalpha.com}
\renewcommand{\shortauthors}{Benhamou et al.}

\begin{abstract}
Financial research agents can generate and monitor trading rules faster than a
human can document the resulting selection process.  Candidates appear at
data-dependent times, weak rules are abandoned, and promising ones are watched
until they cross a reporting threshold.  We introduce \PROMOTE, a protocol that
assigns every candidate predictable testing wealth at birth and accumulates a
cost-aware e-process only from subsequent returns.  If lifetime allocations sum
to at most one, a weighted Ville argument bounds the probability of ever
promoting a rule whose declared bounded net score has always had nonpositive
conditional mean.  The result permits dependent candidates, adaptive births,
and indefinite monitoring.  We audit the implementation in two settings.  A
retrospective pseudo-out-of-sample replay contains \RuleCount\ panel--rule
hypotheses, \NullReps\ shared-sign null paths, injected alternatives, and public
ETF and industry data.  Repeated unadjusted tests produce a family crossing in
\NaiveFWER\ of null paths, versus \SafeFWER\ for \PROMOTE; at a 15-bp aligned
ETF-component perturbation, \PROMOTE\ has \SignalPower\ target power and median
crossing session \MedianDelay.  A separate exact mean-zero experiment lets an
agent select candidate identities and birth times from past scores; the funded
gate has \AdaptiveFWER\ false-family crossings across \AdaptiveReps\ programs.
The market replay produces \RealPromotions\ promotions after costs.  The
certificate controls historical false promotion under the declared null; it is
not a certificate that an earlier edge remains profitable at deployment.

\par\smallskip\noindent\textbf{AI-use disclosure.} Generative AI tools, including OpenAI Codex, assisted manuscript preparation. For this preprint version, Codex assisted with LaTeX editing, author and venue metadata, compilation, and layout checks.
\end{abstract}


\maketitle
\hypersetup{pdfauthor={Eric Benhamou; Jean-Jacques Ohana; Beatrice Guez; Ethan Setrouk; Chamyl Saadi; Thomas Jacquot}}
\begin{center}\small Accepted at ICAIF 2026, Milan, Italy. Author preprint.\end{center}

\section{Introduction}

A conventional backtest has a nominal end date.
Research rarely does.
An analyst looks at interim results, changes the universe, adds a seasonal interaction, drops an unpromising rule, and continues when the evidence is close.
An autonomous research agent can perform the same sequence at machine speed.
Treating the final rule as if it came from one fixed test understates the number of opportunities the system had to be lucky.

This failure is familiar in empirical finance.
Data-snooping corrections compare many strategies on a common terminal sample \cite{white2000reality,hansen2005spa,romano2005stepwise}, while empirical-finance protocols quantify selection and backtest overfitting \cite{harvey2016crosssection,bailey2017pbo,arnott2019protocol}.
Neither device, by itself, authorizes continuous inspection of a library whose membership is decided from the observed history.
The relevant error event is operational: \emph{did the research process ever promote a strategy whose declared net score has no positive conditional mean?}

Test martingales and e-values are well suited to this question because they remain valid at stopping times \cite{ville1939,shafer2011testmartingales,vovk2021evalues,ramdas2023savi}.
Their use still requires a protocol.
A newly proposed rule must not inherit evidence from before it was specified; adding near-duplicates must consume a finite research budget; transaction costs must be known before the return used to update the test; and trading must begin after, rather than on, the boundary-crossing observation.
Those details are where an apparently safe backtest can leak.

\PROMOTE turns them into a ledger.
At a predictable birth time, candidate $h$ receives initial testing wealth $\pi_h$.
Subsequent bounded, cost-adjusted payoffs update an e-process with bets chosen from the past.
Promotion occurs at the first crossing of $1/(\alpha\pi_h)$.
If the total wealth assigned over the life of the research program is at most one, a weighted Ville argument controls the probability of any false promotion by $\alpha$.
The library may be dependent, monitored at different frequencies, and expanded in response to old data.

\paragraph{Contributions.}
The statistical ingredients are established; our contribution is to make them
operational for an adaptive financial-research workflow:
\begin{itemize}
  \setlength{\itemsep}{1pt}
  \setlength{\parskip}{0pt}
  \item a pathwise testing-wealth ledger that covers dependent, predictably
  registered candidates and continuous monitoring;
  \item a cost-aware implementation that separates pre-registration data,
  evidence, promotion, and next-session execution; and
  \item matched fixed-family and adaptive-birth audits that expose false-family
  error, detection delay, and the economic weakness of the historical library.
\end{itemize}
The protocol tests an executable bounded payoff, not a semantic strategy story
supplied by an agent.  Its certificate concerns the complete promotion history,
including rejected ideas; the market replay is a deliberately narrower
pseudo-out-of-sample special case.

\section{Research Process and Null Hypotheses}

Let $(\F_t)_{t\ge0}$ contain all prices, candidate definitions, test states, and decisions available immediately before outcome $t+1$.
A research agent may create candidate $h$ at a stopping time $\nu_h$.
Its rule is an executable map from $\F_{t-1}$ to an exposure $a_{h,t}\in[-1,1]$ for $t>\nu_h$.
The candidate definition includes its asset, direction, rebalance convention, and cost model; changing any of these fields creates a new candidate.

In the ledger, ``hash the rule'' means compute a cryptographic fingerprint of
one canonical, machine-readable package containing the executable code and all
settings that determine its trades.  For example, the package includes the
asset universe, signal formula, parameter values, data fields, calendar,
position sizing, rebalance timing, and cost convention.  A function such as
SHA-256 maps those bytes to a short identifier: identical packages produce the
same identifier, while even a small edit almost certainly produces a different
one.  The hash is not the rule, encryption, or proof that the rule is correct;
the full package is retained.  It is a tamper-evident label that lets an auditor
verify exactly which frozen version was registered before later returns arrived.

Let $r_{h,t}^{\mathrm{net}}$ be the return that candidate $h$ would have earned over period $t$ after its predictable transaction cost.
We use a predictable positive scale $s_{h,t}$ and define
\begin{equation}
Y_{h,t}=\operatorname{clip}\!\left(r_{h,t}^{\mathrm{net}}/s_{h,t},-1,1\right).
\label{eq:score}
\end{equation}
Clipping is part of the estimand: the test concerns the conditional mean of the bounded score, not the unbounded mean return.
Economic results are always computed from the unclipped net return.

For each candidate, the null is
\begin{equation}
H_h:\qquad \E[Y_{h,t}\mid\F_{t-1}]\le0
\quad\text{for every }t>\nu_h.
\label{eq:null}
\end{equation}
This is stronger than a zero unconditional full-sample mean.
It defines a uniform post-birth null: the false-promotion guarantee applies only
when the conditional mean has been nonpositive at every scored date.  If an edge
was positive and later decays, the null is false and old evidence does not
certify that the edge remains positive today.

\begin{assumption}[Predictable registration]
\label{ass:predictable}
The candidate definition, scale $s_{h,t}$, transaction cost, and bet $\lambda_{h,t}$ are $\F_{t-1}$-measurable, with $0\le\lambda_{h,t}\le1$.
The first scored return begins strictly after $\nu_h$.
\end{assumption}

The score may be inactive on dates when the rule holds cash; equivalently, $Y_{h,t}=0$ on those dates.
Candidates can share assets and returns, so no cross-rule independence is assumed.

\section{A Wealth-Conserving Promotion Protocol}

\subsection{One candidate}

Starting from $E_{h,\nu_h}=1$, define
\begin{equation}
E_{h,t}=\prod_{s=\nu_h+1}^{t}
\left(1+\lambda_{h,s}Y_{h,s}\right).
\label{eq:eprocess}
\end{equation}
The nonnegative factor is the return on one unit of statistical wealth bet against $H_h$.

\begin{proposition}[Anytime validity]
\label{prop:anytime}
Under $H_h$ and Assumption~\ref{ass:predictable}, $(E_{h,t})_{t\ge\nu_h}$ is a nonnegative supermartingale.
For every $u\in(0,1)$,
\[
\Pp_{H_h}\!\left(\sup_{t\ge\nu_h}E_{h,t}\ge u^{-1}\right)\le u.
\]
\end{proposition}

\begin{proof}
Conditional on $\F_{t-1}$,
$\E[1+\lambda_{h,t}Y_{h,t}\mid\F_{t-1}]\le1$.
Thus \eqref{eq:eprocess} is a nonnegative supermartingale beginning at one.
Ville's inequality gives the crossing bound.
\end{proof}

The product in \eqref{eq:eprocess} is cumulative.
A rolling product that deletes old factors is generally not an e-process and cannot use Proposition~\ref{prop:anytime}.
Restarting is possible, but every restart must be funded as a new test or combined through a valid mixture.

The implementation avoids choosing a bet after seeing test returns.
For the fixed grid $\Lambda=\{.025,.05,.10,.20,.35\}$, let $E^{(\lambda)}_{h,t}$ denote \eqref{eq:eprocess} with constant bet $\lambda$ and set
\begin{equation}
\bar E_{h,t}=\frac{1}{|\Lambda|}\sum_{\lambda\in\Lambda}E^{(\lambda)}_{h,t}.
\label{eq:mixture}
\end{equation}
Each component starts at one and is a nonnegative supermartingale, so their fixed convex mixture is one as well.
The promotion rule uses $\bar E_{h,t}$ throughout; it never substitutes the best component in hindsight.

\subsection{An adaptively growing library}

Suppose a budget account assigns candidate $h$ a nonnegative mass $\pi_h$ at birth.
The allocation may depend on $\F_{\nu_h}$, but the pathwise total must satisfy
\begin{equation}
\sum_{h\ \mathrm{ever\ created}}\pi_h\le1.
\label{eq:wealth}
\end{equation}
The rule is promoted at
\begin{equation}
\tau_h=\inf\{t>\nu_h:E_{h,t}\ge(\alpha\pi_h)^{-1}\}.
\label{eq:boundary}
\end{equation}

\begin{theorem}[False-promotion control]
\label{thm:fwer}
Under Assumption~\ref{ass:predictable} and \eqref{eq:wealth},
\[
\Pp\{\text{at least one true }H_h\text{ is ever promoted}\}
\le\alpha .
\]
The statement allows dependent candidates, data-dependent birth times, and an unbounded monitoring horizon.
\end{theorem}

\begin{proof}
Condition on $\F_{\nu_h}$.
Proposition~\ref{prop:anytime} bounds the crossing probability of a null candidate by $\alpha\pi_h$.
A union bound over the countable set of created nulls and \eqref{eq:wealth} give the result.
Random predictable masses are handled by iterated expectation.
\end{proof}

Conditioning is what permits adaptive research: pre-birth observations may
determine the rule's code, identity, and mass, but cannot enter its post-birth
product.  Abandonment can set future bets to zero but does not replenish the
account; the program already enjoyed a crossing opportunity.  Revival,
rescaling, or a nearby parameterization requires another debit.  Zero-mass
candidates are ineligible.  These rules block free restarts and hidden
near-duplicate tests.

\begin{figure}[t]
\centering
\fbox{\begin{minipage}{0.94\columnwidth}
\footnotesize
\textbf{\PROMOTE: auditable promotion ledger}
\smallskip

\begin{tabular}{@{}>{\bfseries}r@{\hspace{4pt}}p{0.84\columnwidth}@{}}
1. & Before outcome $t+1$, freeze the rule package, compute its fingerprint,
and record its birth
$\nu_h$, scale, cost convention, bet mixture, and mass $\pi_h$.\\
2. & Debit $\pi_h$ from the remaining lifetime budget; reject a birth that
would make cumulative allocations exceed one.\\
3. & For each $t>\nu_h$, append the realized net score $Y_{h,t}$ and update
every predeclared mixture component.  Never import pre-birth evidence.\\
4. & Record $\bar E_{h,t}$, the applicable boundary
$(\alpha\pi_h)^{-1}$, and the crossing decision.\\
5. & If a crossing occurs at $\tau_h$, schedule economic exposure no earlier
than $\tau_h+1$ and maintain evidence and P\&L as separate ledgers.\\
\end{tabular}
\smallskip
\hrule
\smallskip
\textit{Minimum row key:} candidate hash, event time, data-version hash,
score, cost, evidence, boundary, decision, and first executable session.
\end{minipage}}
\Description{A five-step append-only ledger algorithm: register and hash the rule, debit predictable testing wealth, update only with post-birth scores, record evidence and boundary decisions, and begin economic exposure no earlier than the next session.}
\caption{The protocol is implementable as an append-only ledger.  A changed
rule, restart, or current-regime certificate is a newly funded candidate.}
\label{fig:ledger-algorithm}
\end{figure}

This theorem controls familywise error, a demanding criterion appropriate to an automated promotion gate.
If the objective is discovery rather than certification, stopped e-values can instead be passed to e-BH, which controls FDR under arbitrary dependence \cite{wang2022ebh}.
That is a different operating point and is not used for the primary experiment.

\subsection{Promotion is not same-period trading}

The crossing observation belongs to the test, not to post-promotion performance.
If $\tau_h<\infty$, live or holdout exposure starts at $\tau_h+1$.
This one-period separation prevents the return that triggers promotion from being counted as evidence of deployed profitability.
Statistical wealth is also not trading capital: $E_{h,t}$ may be highly volatile even when the economic position is small.

For a promoted rule, we therefore maintain two ledgers.
The evidence ledger continues to update \eqref{eq:eprocess}; the economic ledger records the actual net return from the next executable interval.
Demotion can be governed by a separately registered rule, but optional removal does not retroactively change the false-promotion guarantee.
Production use therefore requires a distinct current-regime or forward
deployment check.  A predictable epoch restart can provide one, but it must be
funded as a new test; deleting old factors from the original product would break
anytime validity.

\subsection{State, corrections, and recovery}

Each update reads the committed state at $t-1$, joins one point-in-time outcome,
and writes one state for $t$.  Its unique key combines the candidate
fingerprint, event time, and data-version fingerprint.  Replaying a key after a
crash returns the stored row instead of multiplying an e-factor twice; partial
writes are not evidence.

Event time and ingestion time remain separate.  Revised data or code cannot
silently overwrite an old factor: the ledger retains the original row, flags
the certificate, and computes a versioned diagnostic.  Any corrected series
used for authorization starts prospectively as a newly funded stream.  Thus a
reviewer can reconstruct what the gate actually knew when it acted.

Automated checks enforce idempotent replay, no influence from post-$t$ inputs,
and cumulative birth debits that never decrease or exceed one.  These target
double counting, look-ahead joins, and budget recycling.

\subsection{A simple power calculation}

Under an alternative with $\E[Y_{h,t}\mid\F_{t-1}]\ge\mu>0$ and $|Y_{h,t}|\le1$, a fixed bet $\lambda\in(0,1)$ obeys
\begin{equation}
\log(1+\lambda Y_{h,t})
\ge \lambda Y_{h,t}-
\frac{\lambda^2Y_{h,t}^2}{2(1-\lambda)}.
\label{eq:logbound}
\end{equation}
If the predictable second moment is bounded by $v$, the expected log-growth per active observation is at least
$\lambda\mu-\lambda^2v/[2(1-\lambda)]$.
For every component in the implemented grid, \eqref{eq:mixture} gives
$\log \bar E_{h,t}\ge \log E^{(\lambda)}_{h,t}-\log 5$.
Thus adaptation over the five predeclared bet sizes costs at most $\log 5$ units of log evidence relative to the best component on that path; it does not license selecting the best component after the test.
The growth bound makes the cost of a small $\pi_h$ explicit: a candidate born late or deep in a hypothesis tree needs more observations to reach \eqref{eq:boundary}.
Validity is cheap computationally but not statistically.

\section{Experimental Design}

\subsection{Retrospective pseudo-out-of-sample library}

The implementation crosses three asset panels with eight calendar definitions,
giving \RuleCount\ panel--rule hypotheses.  This is the fixed-family special
case of Theorem~\ref{thm:fwer}: within the replay, all candidates are frozen
before the 2015 segment is processed and receive equal wealth.
Adaptive births and data-dependent allocations are not exercised by the market
replay; Section~\ref{sec:adaptive-audit} isolates them under an exact mean-zero
score law.
The panels are six equity ETFs (SPY, QQQ, IWM, EFA, EEM, and VNQ), five defensive or diversifying ETFs (IEF, TLT, GLD, DBC, and SHY), and the ten value-weighted Fama--French industries.
The rules are turn of month, first three and last three trading days, Monday, Friday, last three trading days of a quarter, January, and the third Friday of a month.
The common daily sample runs from March 2006 through May 2026.

For rule indicator $I_{h,t}$, the candidate takes the approximately calendar-neutral panel exposure
\[
q_{h,t}=d_h\frac{I_{h,t}-\hat p_h}{\max(\hat p_h,1-\hat p_h)},
\]
split equally across the assets in its panel.
The activation frequency $\hat p_h$ and direction $d_h\in\{-1,1\}$ are fixed on March 2006--December 2010.
The predictable score scale is the larger of 50 bp and the training 99.5th percentile of absolute net return.
The 2011--2014 segment records stability and component-bet diagnostics, but does not choose the test bet or alter the family.
Every test stream uses the predeclared equal mixture in \eqref{eq:mixture} and restarts at one in January 2015.
The replay code does not use a post-2014 return to change these choices; the
overall split and library were nevertheless designed retrospectively.

The replay segment covers January 2015 through May 2026.
Candidates divide wealth equally, so each boundary is $\RuleCount/\alpha$.
All strategies pay five basis points per dollar of one-way risky-asset turnover.
The code solves the proportional-cost self-financing equation against drifted pre-trade weights rather than subtracting cost from stale targets.
At each date it records exposures and costs, observes the return, updates evidence, and schedules any promotion for the following trading day.

\subsection{Shared-sign null and power experiments}

Type-I error is evaluated on \NullReps\ Monte Carlo paths using a shared-date sign randomization of the observed gross test returns.
The same independent Rademacher sign multiplies the complete cross-hypothesis vector on a date, preserving observed magnitudes and contemporaneous cross-rule dependence while removing serial sign dependence.
The nonnegative cost-fraction path is conditioned upon rather than recomputed, and scores are recalculated and clipped after costs.
This construction satisfies \eqref{eq:null} conditionally on the magnitudes and costs.
Indeed, with $f(u)=\operatorname{clip}(u,-1,1)$, $a=-c/s\le0$, $b=(1-c)g/s$, and a Rademacher sign $S$,
\[
\E_S f(a+Sb)=\tfrac12\{f(a+b)+f(a-b)\}
\le\tfrac12\{f(b)+f(-b)\}=0,
\]
because $f$ is increasing and odd.
The complete fixed-family monitoring rule is rerun, including every intermediate inspection.
The primary statistic is whether at least one null candidate is ever promoted.

The comparator repeatedly computes an expanding one-sided z statistic after a minimum of 63 sessions and declares success whenever an unadjusted $p$-value falls below $\alpha$ at an inspection date.
This baseline is intentionally common rather than defensible: it measures the inflation caused by the workflow that \PROMOTE is designed to replace.
A second comparator gives every stream its own anytime boundary $1/\alpha$ without splitting wealth; it isolates cross-candidate multiplicity from optional stopping.
A matched-error fixed-horizon comparator rejects candidate $h$ only when its terminal mixture satisfies
$\bar E_{h,n}\ge(\alpha\pi_h)^{-1}$.
Markov's inequality and a union bound give the same familywise guarantee under \eqref{eq:null}; unlike \PROMOTE, this comparator discards earlier crossings and cannot stop before session $n$.

Power paths shift each equity component return on every date by 5, 10, 15, 20, or 30 basis points in the sign of the target equity turn-of-month weight.
The target therefore receives a positive gross contribution proportional to its absolute exposure, and overlapping equity rules also move.
The five shifts correspond to 4.8\%, 9.6\%, 14.4\%, 19.2\%, and 28.8\% annualized gross alpha for the target overlay.
We report target detection and delay; the repeated-$p$ curve is diagnostic and is not a matched-Type-I-error comparison.

\subsection{Adaptive-birth boundary-null audit}
\label{sec:adaptive-audit}

The fixed market replay cannot demonstrate the adaptive-birth part of
Theorem~\ref{thm:fwer}.  We therefore add a separate exact boundary-null
experiment.  Each of \AdaptiveReps\ research programs contains 32 independent
Rademacher score streams over 768 sessions.  After a 64-session warm-up, an
agent inspects the past only and registers the unregistered stream with the
largest trailing-64 sum whenever that sum is at least 20.  Births are separated
by 32 sessions and capped at 12.  The selected identity and birth time are thus
data dependent, but its first test score is the next independent Rademacher
increment, so
\(\E[Y_{h,t}\mid\F_{t-1}]=0\) exactly after registration.

At each birth the agent predictably spends one half of remaining wealth when
the trailing sum is at least 28 and one quarter otherwise.  Consequently the
pathwise lifetime allocation never exceeds one even though candidate identity,
birth time, mass, and final family size vary by path.  We compare the funded
boundary with an unfunded per-stream boundary of 20 and with repeated one-sided
Gaussian tests after 32 post-birth observations.  The simulation uses seed
20262731 and the same fixed e-mixture as the market study.

\subsection{Public-market replay}

The market study uses a July 31, 2026 frozen snapshot: Yahoo adjusted closes for
the ETFs and the Ken French Data Library's value-weighted ten-industry file.
Source and derived-file SHA-256 hashes accompany the artifact; the replay itself
performs no network retrieval.
Returns, candidate indicators, costs, and evidence are computed from information available at each decision.
The study reports the entire promotion ledger, including candidates that never cross.
The split and library were designed retrospectively, and trading-day ranks are reconstructed from the realized exchange calendar.
No test return calibrates direction, scale, bet grid, or prior wealth, but one historical path cannot establish the conditional null in \eqref{eq:null}.

For any promoted strategy, exposure begins one interval after crossing and pays exact one-way turnover under the declared cost convention.
Program-level wealth includes cash before activation; the complete daily ledger permits a separate live-only analysis when a crossing occurs.

\section{Results}

\subsection{Fixed-family calibration and power}

\begin{table}[t]
\caption{Fixed-family audit. Entries are percentages [95\% Wilson interval].
Null rates use \NullReps\ paths; 15-bp target power uses 1,500.  Median is the
one-based crossing session conditional on detection.}
\label{tab:simulation}
\centering
\scriptsize
\setlength{\tabcolsep}{2.0pt}
\begin{tabular}{@{}lrrr@{}}
\toprule
Procedure & Null FWER & 15-bp power & Med. session\\
\midrule
Repeated $p<.05$ & 94.5 [93.6,95.3] & 99.9 [99.6,100.0] & 125\\
Fixed $\bar E_n\ge480$ & 0.0 [0,.128] & 67.1 [64.7,69.5] & 2,868\\
Per-stream $\bar E_t\ge20$ & 0.7 [.46,1.07] & --- & ---\\
\PROMOTE\ $\bar E_t\ge480$ & 0.0 [0,.128] & 70.1 [67.7,72.3] & 2,327\\
\bottomrule
\end{tabular}
\end{table}

Table~\ref{tab:simulation} gives the principal calibration result.
Continuous use of unadjusted fixed-sample tests turns a nominal threshold into a much larger research-program error rate.
The wealth-split rule records no false family crossing in 3,000 paths; its
Wilson upper endpoint is 0.128\%.  This shared-sign law includes frozen positive
costs and is therefore generally a strict, conservative null rather than a
least-favorable boundary case.  Giving every stream its own boundary of 20 has
no 5\% familywise guarantee; it records 21 crossings (0.7\%) here, still below
5\% because the component e-processes are conservative under this law.  The
fixed-horizon and anytime wealth-split rules both record zero crossings.

The injected alternatives show the price of validity.
Evidence accumulates only on observations realized after the test starts, and a candidate with mass $\pi_h$ must earn $\log(1/(\alpha\pi_h))$ units of log evidence.
Strong, persistent effects are promoted; weak or intermittent effects often remain below the boundary during the available horizon.
Target power rises from 0\% at 5 bp to 7.7\%, 70.1\%, 99.3\%, and 100\% at 10, 15, 20, and 30 bp, respectively.
At 15 bp, the matched terminal rule has 67.1\% power (95\% Wilson interval 64.7--69.5\%), while \PROMOTE has 70.1\% (67.7--72.3\%) and crosses at a median 2,327 of 2,868 sessions.
The terminal rule necessarily waits for session 2,868; the naive test crosses after 125 sessions in paths where it detects the target, but its null FWER is 94.5\%.
At 15 bp the anytime event contains 3.0 percentage points of paths that cross
before the horizon but finish below the terminal boundary.  This is useful for a
gate, although \PROMOTE\ is not designed to maximize discovery count.

\begin{table}[t]
\caption{Complete target-power frontier (1,500 paths per shift).  Shift is bp
per ETF component per session, with annualized target gross alpha in
parentheses.  Power entries are percentages [95\% Wilson interval]; $t_{50}$ is
the anytime median crossing session conditional on detection.  A terminal
detection, when present, occurs at session 2,868.}
\label{tab:power-frontier}
\centering
\scriptsize
\setlength{\tabcolsep}{2.0pt}
\begin{tabular}{@{}rrrr@{}}
\toprule
Shift (ann. \%) & Terminal power & Anytime power & $t_{50}$\\
\midrule
0 (0.0)   & 0.00 [0,.26]       & 0.00 [0,.26]       & ---\\
5 (4.8)   & 0.00 [0,.26]       & 0.00 [0,.26]       & ---\\
10 (9.6)  & 5.73 [4.67,7.03]   & 7.73 [6.49,9.20]   & 2,568\\
15 (14.4) & 67.13 [64.72,69.46]& 70.07 [67.70,72.33]& 2,327\\
20 (19.2) & 99.20 [98.61,99.54]& 99.27 [98.69,99.59]& 1,743\\
30 (28.8) & 100 [99.74,100]    & 100 [99.74,100]    & 1,049\\
\bottomrule
\end{tabular}
\end{table}

Table~\ref{tab:power-frontier} exposes the full operating curve rather than a
single favorable effect.  The valid procedures detect none of the zero- or
5-bp target paths; with 1,500 paths, the zero-count Wilson upper endpoint is
0.26\%.  The repeated z workflow, by comparison, declares the target on 23.2\%
of zero-shift paths and 74.2\% of 5-bp paths, so its apparent sensitivity cannot
be separated from its inflated inspection error.  Between 10 and 15 bp the
funded methods move through their steep transition.  The anytime rule gains
2.0 and 2.9 power points over the terminal rule, respectively, because it
retains paths that cross earlier and later fall below 480; overlapping
intervals caution against interpreting these differences as a tuned advantage.
At 20 bp the power gap nearly disappears, but the anytime median crossing is
1,125 sessions earlier than the terminal decision.  Even at 30 bp, its median
requires 1,049 sessions.  The boundary therefore rewards persistent net
evidence, not a visually strong but short backtest.

\begin{figure*}[!t]
\centering
\includegraphics[width=\textwidth]{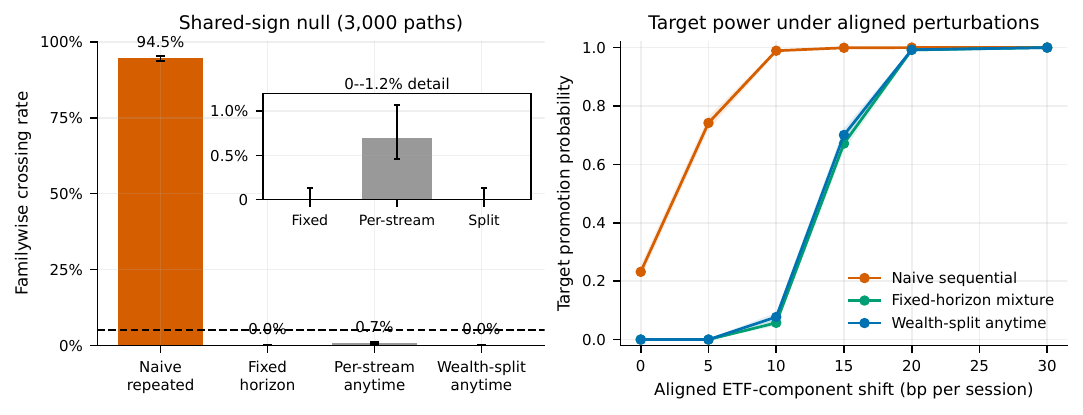}
\Description{Two diagnostics: full-scale null familywise crossing rates with a zoomed inset for rates below 1.2 percent, and target-power curves with Wilson bands over aligned ETF-component perturbations.}
\caption{Fixed-family diagnostics.  Left: familywise crossings over candidates
and inspection times; the inset makes the 0.7\% per-stream result and zero-count
Wilson intervals visible.  Right: target promotion probability under an aligned
shift to each ETF component return; shaded regions are 95\% Wilson intervals.
The repeated-test curve is diagnostic, not a matched-error comparator.}
\label{fig:diagnostics}
\end{figure*}

\subsection{Adaptive registration on an exact boundary null}

\begin{table}[t]
\caption{Adaptive-birth mean-zero audit (\AdaptiveReps\ programs).  Intervals
are 95\% Wilson intervals in percentage points.}
\label{tab:adaptive}
\centering
\small
\setlength{\tabcolsep}{3.0pt}
\begin{tabular}{lrr}
\toprule
Procedure & Crossings & FWER [95\% CI]\\
\midrule
Repeated one-sided $p<.05$ & 2,792 & 93.07 [92.10, 93.92]\\
Anytime per stream, $\bar E_t\ge20$ & 552 & 18.40 [17.05, 19.83]\\
Funded adaptive \PROMOTE & 53 & 1.77 [1.35, 2.30]\\
\bottomrule
\end{tabular}
\end{table}

The agent registers a mean of \AdaptiveBirths\ candidates per program (median
10; 10th--90th percentiles 8--12) after selecting strong pre-birth histories.
It allocates 93.45\% of testing wealth on average, while the pathwise total never
exceeds one.  Because only independent future increments enter evidence, this
experiment sits exactly on the conditional mean-zero boundary.  Table
\ref{tab:adaptive} shows that adaptive selection itself does not invalidate the
funded gate: its interval lies below 5\%.  In contrast, reusing a full 5\%
threshold for each adaptively selected stream produces 18.4\% familywise error.

\begin{table}[t]
\caption{Predictable allocation sensitivity on the same 3,000 adaptive null
programs.  ``First'' and ``spent'' are mean percentages of initial testing
wealth; FWER is a percentage [95\% Wilson interval].}
\label{tab:allocation-sensitivity}
\centering
\scriptsize
\setlength{\tabcolsep}{2.5pt}
\begin{tabular}{@{}lrrr@{}}
\toprule
Allocation rule & First & Spent & FWER [95\% CI]\\
\midrule
Observed-strength split & 25.29 & 93.45 & 1.77 [1.35,2.30]\\
Equal 12-slot reserve & 8.33 & 82.36 & 1.17 [0.84,1.62]\\
Quarter remaining & 25.00 & 93.34 & 1.77 [1.35,2.30]\\
Half remaining & 50.00 & 99.75 & 2.37 [1.88,2.97]\\
\bottomrule
\end{tabular}
\end{table}

Table~\ref{tab:allocation-sensitivity} replays the identical births and
post-birth evidence under four predeclared spending policies.  The equal-slot
rule reserves $1/12$ for every possible birth.  The geometric rules spend a
quarter or a half of remaining wealth, and the observed-strength rule uses the
pre-birth trailing sum to choose between those fractions.  Each policy obeys
$\sum_h\pi_h\le1$ pathwise; unfilled equal slots and the tail of a geometric
schedule remain unspent.  Thus Theorem~\ref{thm:fwer} applies to all four
without assuming how many candidates will arrive.

The estimated FWER stays below 5\% for every policy, but the finite-sample
rates illustrate a design trade-off rather than a ranking theorem.  Spending
half of the remainder gives early proposals the lowest boundary, then rapidly
shrinks later masses; it also has the largest crossing rate here.  Equal slots
retain more capacity for late births and spend only 82.36\% on average because
programs register fewer than 12 candidates.  The signal-dependent and
quarter-remaining rules happen to produce the same 53 crossing programs, but
their masses are not generally identical.  Allocation may favor proposals
using information available at birth; it must not be revised after observing
their test evidence.  This separates an auditable research priority from
post-selection threshold shopping.

\subsection{Market evidence and the finance ledger}

The pseudo-out-of-sample market run promotes \RealPromotions\ of \RuleCount\ candidates.
This count should not be read as a rejection rate under a known stochastic null: market returns need not satisfy \eqref{eq:null}, and the experiment supplies only one dependent path.
Its role is to show how the protocol behaves when familiar strategy families compete for the same research budget.

The largest evidence value reached by any candidate is 1.93, versus the declared boundary of 480.
Repeated unadjusted inspection instead promotes the reversed Fama--French industry January rule on 13 January 2016.
Its full 2015--2026 shadow path loses 0.90\% per year (Sharpe $-0.18$); the fixed $1/24$ sleeve activated by the naive gate ends at 0.992 after costs.
These figures do not prove that the crossing was a statistical false positive, but they show why a nominal $p<.05$ crossing is a poor deployment event.

Only two of the 24 full-test shadow paths have a positive arithmetic mean, and the larger is 0.087\% annualized; neither comes close to the boundary.
This is a weak library after the retrospective split, not evidence that the gate suppresses profitable rules.
Conversely, any promoted rule has an audit trail identifying the exact observations, score scaling, cost, bet, and boundary responsible for the decision.
This is more informative than reporting only the winning backtest.

\begin{figure*}[!t]
\centering
\includegraphics[width=\textwidth]{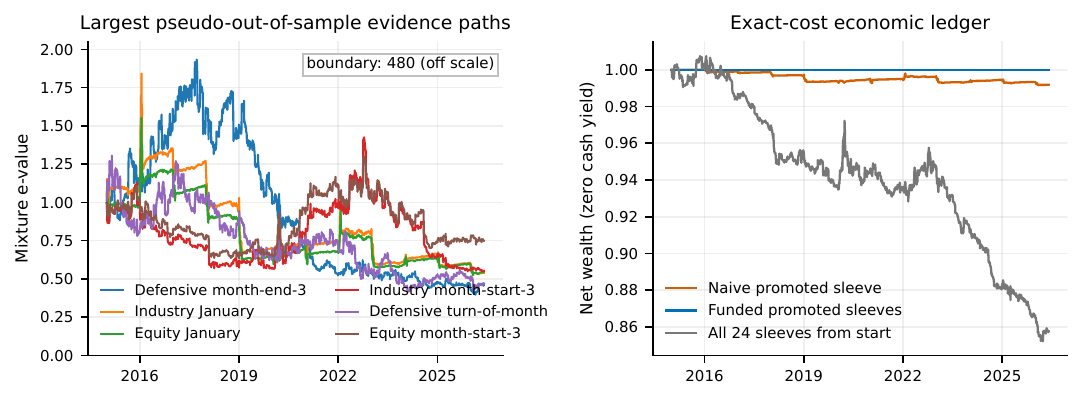}
\Description{Two market-replay panels: six readable historical mixture e-value paths on a zero-to-two scale with the boundary stated as 480 off scale, and exact-cost wealth for the naive promoted sleeve, funded promoted sleeves, and all 24 sleeves from inception.}
\caption{Pseudo-out-of-sample evidence and economic ledgers, January
2015--May 2026.  Left: the six largest mixture e-processes; the artifact retains
all 24 paths.  Right: portfolio wealth after exact five-basis-point one-way
turnover costs.  The funded gate stays at one because no rule crosses; cash is
assigned zero yield in this diagnostic ledger.}
\label{fig:market-ledger}
\end{figure*}

\begin{table}[t]
\caption{Finance audit from the exact daily ledger. Returns, volatility, and
one-way risky-asset turnover are annualized; SR uses zero cash yield. All paths
pay 5 bp per dollar traded. Industry sleeves are non-tradable, so the economics
are diagnostic.}
\label{tab:finance-audit}
\centering
\scriptsize
\setlength{\tabcolsep}{3.0pt}
\begin{tabular}{lrrr}
\toprule
Metric & Funded \PROMOTE & Naive sleeve & All 24 sleeves\\
\midrule
Annualized net return & 0.000\% & $-0.071\%$ & $-1.341\%$\\
Annualized volatility & 0.000\% & 0.167\% & 1.577\%\\
Sharpe, zero cash yield & 0.00 & $-0.43$ & $-0.85$\\
Maximum drawdown & 0.00\% & $-0.90\%$ & $-15.41\%$\\
One-way turnover/year & 0.00 & 0.10 & 26.06\\
Mean gross exposure & 0.00\% & 0.55\% & 10.23\%\\
Terminal wealth & 1.000 & 0.992 & 0.858\\
\bottomrule
\end{tabular}
\end{table}

\subsection{What the artifacts establish}

The rowwise replay verifies three deterministic invariants: the 24 fixed-family
weights sum to one, every test e-process restarts at one, and no promoted return
is credited before the interval following a crossing.  The adaptive simulation
checks variable birth times, identities, masses, and family sizes under an exact
mean-zero law.  Seeds 20260731 and 20262731, the complete hypothesis ledger,
summary tables for 15,000 simulated programs or paths, and SHA-256 hashes of
every machine-readable result accompany the artifact.

These checks do not establish a stable positive conditional mean for a market
rule.  Power depends on clipping, effect persistence, family size, and bet
fractions.  Changing those choices after seeing the replay creates another
research layer that requires new validation data or testing wealth.

\FloatBarrier

\section{Relation to Existing Work}

Backtest overfitting has been studied through data-snooping tests, superior-predictive-ability tests, multiple-testing adjustments, and selection diagnostics \cite{white2000reality,hansen2005spa,romano2005stepwise,harvey2016crosssection,bailey2017pbo}.
These methods address closely related selection problems, usually for a collection evaluated at a terminal sample size.
Our target is the event generated by an ongoing research system: a false strategy is promoted at some random time after a data-dependent proposal.

Repeated interaction with a holdout set is adaptive data analysis; the
reusable-holdout framework gives a privacy-based route to valid reuse
\cite{dwork2015adaptive}.  Asynchronous online testing handles overlapping
sequential experiments \cite{zrnic2021asynchronous}.  Neither is a cost-aware
familywise promotion ledger for financial strategies.

Online multiple-testing methods allocate error across a sequence of hypotheses,
including alpha-investing and adaptive FDR rules
\cite{foster2008alphainvesting,javanmard2018onlinefdr,ramdas2018saffron}.
Our pathwise masses are deliberately simpler: they are a weighted online
Bonferroni allocation for familywise control, with no claim of wealth recycling.
That conservative choice makes the audit rule transparent when candidate
identities and birth times are selected from the past.

E-values provide the underlying evidence currency \cite{vovk2021evalues}, and safe testing, betting-based inference, and confidence sequences explain their optional-stopping behavior \cite{grunwald2024safetesting,waudbysmith2024betting,howard2021confidence,ramdas2023savi}.
The e-BH procedure extends the framework to false-discovery control under arbitrary dependence \cite{wang2022ebh}.
Sequential e-values have also been used to evaluate probability forecasts and backtest financial risk measures \cite{henzi2022sequential,wang2025ebacktesting}; those applications test forecasts, whereas our unit is a strategy-promotion workflow.
We use the conservative wealth-weighted crossing rule because one false launch
can matter operationally.  Our contribution is its implementation as a
strategy-level gate, an exact adaptive-birth audit, and a reproducible market
ledger with costs and delayed deployment.

Calendar and technical trading rules provide a demanding case study because their design space is easy to expand and many reported effects are sample-sensitive \cite{lakonishok1988seasonal,sullivan1999technical,sullivan2001calendar,bajgrowicz2012technical}.
More broadly, the attenuation of published return predictability and the prevalence of false anomaly rejections motivate a strict separation between discovery and deployment \cite{mclean2016predictability,chordia2020anomalies}.
The candidate generator may instead be a language model, feature search, or a
human, provided its executable definition and wealth are fixed before scoring.
ICAIF studies multiscale evaluation of financial time-series foundation models
\cite{cheong2025bfortfin}, and AAAI studies trading-policy generalization beyond
offline data \cite{yuan2026metatrader}.  Neither controls the probability of
ever promoting a null strategy during adaptive search.

\subsection{Expiring evidence and predictably funded revalidation}
\label{sec:epoch-revalidation}

A promotion is historical evidence, not a perpetual certificate of current
profitability.  The process in \eqref{eq:eprocess} is cumulative: evidence
earned during an early favorable regime is not erased when the conditional mean
later becomes zero or negative.  Theorem~\ref{thm:fwer} controls crossing under
the uniformly nonpositive null \eqref{eq:null}; it does not assert that the
conditional mean remains positive after a genuine crossing.  Simply deleting
old factors or resetting the displayed process to one while reusing its original
testing mass would create additional crossing opportunities outside the theorem.

A production protocol can instead make authorization expire and fund every
revalidation as a new candidate.  Give strategy definition $h$ a lifetime base
mass $\omega_h$, allocated predictably at its original birth, with
$\sum_{h\ {\rm ever\ created}}\omega_h\le1$ pathwise.  Choose
$(q_e)_{e\ge0}$ nonnegative with $\sum_e q_e\le1$.  Epoch $e$ may begin at any
stopping time $\rho_{h,e}$ determined from past information and receives fresh
mass $\pi_{h,e}=\omega_hq_e$.  Its e-process starts from one, uses only scores
strictly after $\rho_{h,e}$, and crosses at
$\bar E_{h,e,t}\ge(\alpha\omega_hq_e)^{-1}$.  A finite authorization window
can be declared at epoch birth; setting bets to zero after expiry freezes the
process, and renewal requires a separately funded epoch.

\begin{proposition}[False authorization under funded restarts]
\label{prop:funded-restarts}
Suppose every epoch satisfies Assumption~\ref{ass:predictable}, and define its
null by $\E[Y_{h,t}\mid\F_{t-1}]\le0$ at every time on which that epoch places a
nonzero bet.  If authorization or renewal occurs only after its funded boundary,
then the probability that any null epoch ever authorizes a strategy is at most
$\alpha$.  The result allows data-dependent restart times, overlapping epochs,
shared returns, and an unbounded number of reviews.
\end{proposition}

\begin{proof}
Treat each pair $(h,e)$ as a distinct candidate in
Theorem~\ref{thm:fwer}.  Its predictable mass is $\omega_hq_e$. Moreover,
\[\sum_{h,e}\omega_hq_e\le(\sum_h\omega_h)(\sum_e q_e)\le1.\]
The theorem
applies directly; no independence across epochs is required.
\end{proof}

With $H$ initially equal-weighted definitions, $\omega_h=1/H$, the epoch-$e$
threshold and log-evidence requirement are $H/(\alpha q_e)$ and
$\log(H/\alpha)+\log(1/q_e)$.  Doubling the family adds $\log2$ to the evidence
requirement, while later restarts pay $\log(1/q_e)$.  For example,
$q_e=6/[\pi^2(e+1)^2]$ reserves finite lifetime wealth while making the
threshold grow quadratically rather than permitting cost-free restarts.
Epoching does not prove that a profitable regime persists after crossing; it
makes every renewed authorization depend on fresh post-restart observations
and a declared share of the lifetime error budget.

\section{Limitations and Governance}

The guarantee is narrow by design.  It concerns the clipped, cost-adjusted score
and a null that is uniformly nonpositive after birth; it does not certify that a
formerly positive edge remains profitable.  Clipping can also suppress rare
economic losses.  It occurs in 128 of 68,832 historical candidate--session
scores (0.186\%), so every evidence row is retained beside its unclipped return.

The two simulations answer different questions.  Shared independent signs
preserve magnitudes and contemporaneous cross-rule dependence, but remove
serial sign dependence and freeze observed cost fractions; the resulting null
is generally strict.  The adaptive audit is exactly mean zero and exercises
selection and staggered births, but its iid Rademacher scores are not a market
model.  Neither construction asserts that realized returns satisfy
\eqref{eq:null}.

Validity is only as trustworthy as the filtration.  Revised data,
survivorship-biased universes, pretrained-model knowledge of the replay era, or
same-close execution can import future information.  A production agent must
therefore retain data and prompt hashes, candidate births, allocations, and
rejected proposals.  An idea without a deterministic executable rule receives
no testing wealth.

The economic replay uses adjusted closes and zero cash yield.  Five-basis-point
turnover costs are explicit, but financing, borrow fees, bid--ask dynamics,
impact, taxes, and capacity are absent; French industries are not directly
tradable.  The design and pseudo-out-of-sample periods were chosen
retrospectively, and realized trading-day ranks can encode rare unscheduled
closures.  These facts make Table~\ref{tab:finance-audit} an audit, not a live
performance claim.

Finally, familywise control becomes conservative when an agent proposes many
correlated variants.  Hierarchical predictable allocations expose but do not
eliminate that cost; discovery-oriented systems may prefer FDR control.  In all
cases, promotion should trigger a separately governed forward evaluation rather
than unrestricted scaling.

In production, a crossing is an authorization request, not an order.  A
pre-trade gate still checks data availability, venue and borrow eligibility,
spread and impact, concentration, turnover, and duplicate portfolio exposure.
It may veto or cap a trade without changing the evidence.  A predictable risk
kill switch may stop exposure or betting, but cannot reclaim testing mass or
justify a same-day restart.

The ledger should fail closed: missing prices, ambiguous calendars, stale
features, or a code-hash mismatch suspend updates and deployment until an
operator records the resolution and first usable session.  The e-process asks
whether the historical null was crossed; operations controls ask whether the
position is safe to trade now.

\subsection{Certificate lifecycle under failure}

A certificate moves through registered, accumulating, crossed, authorized,
suspended, and expired states.  Crossing concerns one frozen version;
authorization is separate and time limited.  Expiry requires a funded epoch
from Section~\ref{sec:epoch-revalidation}; corrected history cannot backdate
promotion.

The release manifest binds the fingerprint to its rule package, data vintage,
calendar, costs, bet grid, budget, and expiry.  This supports audits of replay,
post-decision inputs, and the first trade.  It does not prove profitability; it
prevents implementation changes from voiding the theorem's premises.

\section{Conclusion}

Adaptive research changes the unit of inference.
The relevant object is the sequence of proposals, inspections, and stopping
decisions, not one terminal backtest.  \PROMOTE gives that process a finite
error budget: candidates inherit no evidence and must cross a cost-aware,
wealth-adjusted boundary.  The guarantee survives continuous monitoring and an
adaptively expanding library.

The experiments expose the price: repeated inspection finds winners readily,
whereas the valid gate may wait for weak effects.  That delay is the cost of an
optional-stopping guarantee, not evidence of current profitability.  Frictions,
scale, and forward risk remain deployment questions.

\bibliographystyle{ACM-Reference-Format}
\bibliography{assets/2636a5a6_references}


\begin{thebibliography}{29}


\ifx \showCODEN    \undefined \def \showCODEN     #1{\unskip}     \fi
\ifx \showISBNx    \undefined \def \showISBNx     #1{\unskip}     \fi
\ifx \showISBNxiii \undefined \def \showISBNxiii  #1{\unskip}     \fi
\ifx \showISSN     \undefined \def \showISSN      #1{\unskip}     \fi
\ifx \showLCCN     \undefined \def \showLCCN      #1{\unskip}     \fi
\ifx \shownote     \undefined \def \shownote      #1{#1}          \fi
\ifx \showarticletitle \undefined \def \showarticletitle #1{#1}   \fi
\ifx \showURL      \undefined \def \showURL       {\relax}        \fi
\providecommand\bibfield[2]{#2}
\providecommand\bibinfo[2]{#2}
\providecommand\natexlab[1]{#1}
\providecommand\showeprint[2][]{arXiv:#2}

\bibitem[Arnott et~al\mbox{.}(2019)]%
        {arnott2019protocol}
\bibfield{author}{\bibinfo{person}{Robert~D. Arnott},
  \bibinfo{person}{Campbell~R. Harvey}, {and} \bibinfo{person}{Harry
  Markowitz}.} \bibinfo{year}{2019}\natexlab{}.
\newblock \showarticletitle{A Backtesting Protocol in the Era of Machine
  Learning}.
\newblock \bibinfo{journal}{\emph{The Journal of Financial Data Science}}
  \bibinfo{volume}{1}, \bibinfo{number}{1} (\bibinfo{year}{2019}),
  \bibinfo{pages}{64--74}.
\newblock
\href{https://doi.org/10.3905/jfds.2019.1.064}{doi:\nolinkurl{10.3905/jfds.2019.1.064}}


\bibitem[Bailey et~al\mbox{.}(2017)]%
        {bailey2017pbo}
\bibfield{author}{\bibinfo{person}{David~H. Bailey},
  \bibinfo{person}{Jonathan~M. Borwein}, \bibinfo{person}{Marcos L{\'o}pez~de
  Prado}, {and} \bibinfo{person}{Qiji~Jim Zhu}.}
  \bibinfo{year}{2017}\natexlab{}.
\newblock \showarticletitle{The Probability of Backtest Overfitting}.
\newblock \bibinfo{journal}{\emph{The Journal of Computational Finance}}
  \bibinfo{volume}{20}, \bibinfo{number}{4} (\bibinfo{year}{2017}),
  \bibinfo{pages}{39--69}.
\newblock
\href{https://doi.org/10.21314/JCF.2016.322}{doi:\nolinkurl{10.21314/JCF.2016.322}}


\bibitem[Bajgrowicz and Scaillet(2012)]%
        {bajgrowicz2012technical}
\bibfield{author}{\bibinfo{person}{Pierre Bajgrowicz} {and}
  \bibinfo{person}{Olivier Scaillet}.} \bibinfo{year}{2012}\natexlab{}.
\newblock \showarticletitle{Technical Trading Revisited: False Discoveries,
  Persistence Tests, and Transaction Costs}.
\newblock \bibinfo{journal}{\emph{Journal of Financial Economics}}
  \bibinfo{volume}{106}, \bibinfo{number}{3} (\bibinfo{year}{2012}),
  \bibinfo{pages}{473--491}.
\newblock
\href{https://doi.org/10.1016/j.jfineco.2012.06.001}{doi:\nolinkurl{10.1016/j.jfineco.2012.06.001}}


\bibitem[Cheong et~al\mbox{.}(2025)]%
        {cheong2025bfortfin}
\bibfield{author}{\bibinfo{person}{Nigel Cheong}, \bibinfo{person}{Ling~Wei
  Hsuen}, \bibinfo{person}{Ranjan Satapathy}, \bibinfo{person}{Erik Cambria},
  \bibinfo{person}{Rick Siow~Mong Goh}, {and} \bibinfo{person}{Joyjit
  Chattoraj}.} \bibinfo{year}{2025}\natexlab{}.
\newblock \showarticletitle{{BForTFin}: A Financial Domain-Aware Multiscale
  Evaluation Method for Time-Series Foundation Models}. In
  \bibinfo{booktitle}{\emph{Proceedings of the 6th ACM International Conference
  on AI in Finance}}. \bibinfo{publisher}{Association for Computing Machinery},
  \bibinfo{pages}{274--282}.
\newblock
\href{https://doi.org/10.1145/3768292.3770402}{doi:\nolinkurl{10.1145/3768292.3770402}}


\bibitem[Chordia et~al\mbox{.}(2020)]%
        {chordia2020anomalies}
\bibfield{author}{\bibinfo{person}{Tarun Chordia}, \bibinfo{person}{Amit
  Goyal}, {and} \bibinfo{person}{Alessio Saretto}.}
  \bibinfo{year}{2020}\natexlab{}.
\newblock \showarticletitle{Anomalies and False Rejections}.
\newblock \bibinfo{journal}{\emph{The Review of Financial Studies}}
  \bibinfo{volume}{33}, \bibinfo{number}{5} (\bibinfo{year}{2020}),
  \bibinfo{pages}{2134--2179}.
\newblock
\href{https://doi.org/10.1093/rfs/hhaa018}{doi:\nolinkurl{10.1093/rfs/hhaa018}}


\bibitem[Dwork et~al\mbox{.}(2015)]%
        {dwork2015adaptive}
\bibfield{author}{\bibinfo{person}{Cynthia Dwork}, \bibinfo{person}{Vitaly
  Feldman}, \bibinfo{person}{Moritz Hardt}, \bibinfo{person}{Toni Pitassi},
  \bibinfo{person}{Omer Reingold}, {and} \bibinfo{person}{Aaron Roth}.}
  \bibinfo{year}{2015}\natexlab{}.
\newblock \showarticletitle{Generalization in Adaptive Data Analysis and
  Holdout Reuse}. In \bibinfo{booktitle}{\emph{Advances in Neural Information
  Processing Systems 28}}. \bibinfo{publisher}{Curran Associates, Inc.}
\newblock
\urldef\tempurl%
\url{https://papers.nips.cc/paper/5993-generalization-in-adaptive-data-analysis-and-holdout-reuse}
\showURL{%
\tempurl}


\bibitem[Foster and Stine(2008)]%
        {foster2008alphainvesting}
\bibfield{author}{\bibinfo{person}{Dean~P. Foster} {and}
  \bibinfo{person}{Robert~A. Stine}.} \bibinfo{year}{2008}\natexlab{}.
\newblock \showarticletitle{Alpha-Investing: A Procedure for Sequential Control
  of Expected False Discoveries}.
\newblock \bibinfo{journal}{\emph{Journal of the Royal Statistical Society:
  Series B (Statistical Methodology)}} \bibinfo{volume}{70},
  \bibinfo{number}{2} (\bibinfo{year}{2008}), \bibinfo{pages}{429--444}.
\newblock
\href{https://doi.org/10.1111/j.1467-9868.2007.00643.x}{doi:\nolinkurl{10.1111/j.1467-9868.2007.00643.x}}


\bibitem[Gr{\"u}nwald et~al\mbox{.}(2024)]%
        {grunwald2024safetesting}
\bibfield{author}{\bibinfo{person}{Peter Gr{\"u}nwald}, \bibinfo{person}{Rianne
  de Heide}, {and} \bibinfo{person}{Wouter~M. Koolen}.}
  \bibinfo{year}{2024}\natexlab{}.
\newblock \showarticletitle{Safe Testing}.
\newblock \bibinfo{journal}{\emph{Journal of the Royal Statistical Society
  Series B: Statistical Methodology}} \bibinfo{volume}{86}, \bibinfo{number}{5}
  (\bibinfo{year}{2024}), \bibinfo{pages}{1091--1128}.
\newblock
\href{https://doi.org/10.1093/jrsssb/qkae011}{doi:\nolinkurl{10.1093/jrsssb/qkae011}}


\bibitem[Hansen(2005)]%
        {hansen2005spa}
\bibfield{author}{\bibinfo{person}{Peter~Reinhard Hansen}.}
  \bibinfo{year}{2005}\natexlab{}.
\newblock \showarticletitle{A Test for Superior Predictive Ability}.
\newblock \bibinfo{journal}{\emph{Journal of Business \& Economic Statistics}}
  \bibinfo{volume}{23}, \bibinfo{number}{4} (\bibinfo{year}{2005}),
  \bibinfo{pages}{365--380}.
\newblock


\bibitem[Harvey et~al\mbox{.}(2016)]%
        {harvey2016crosssection}
\bibfield{author}{\bibinfo{person}{Campbell~R. Harvey}, \bibinfo{person}{Yan
  Liu}, {and} \bibinfo{person}{Heqing Zhu}.} \bibinfo{year}{2016}\natexlab{}.
\newblock \showarticletitle{{\ldots} and the Cross-Section of Expected
  Returns}.
\newblock \bibinfo{journal}{\emph{The Review of Financial Studies}}
  \bibinfo{volume}{29}, \bibinfo{number}{1} (\bibinfo{year}{2016}),
  \bibinfo{pages}{5--68}.
\newblock
\href{https://doi.org/10.1093/rfs/hhv059}{doi:\nolinkurl{10.1093/rfs/hhv059}}


\bibitem[Henzi and Ziegel(2022)]%
        {henzi2022sequential}
\bibfield{author}{\bibinfo{person}{Alexander Henzi} {and}
  \bibinfo{person}{Johanna~F. Ziegel}.} \bibinfo{year}{2022}\natexlab{}.
\newblock \showarticletitle{Valid Sequential Inference on Probability Forecast
  Performance}.
\newblock \bibinfo{journal}{\emph{Biometrika}} \bibinfo{volume}{109},
  \bibinfo{number}{3} (\bibinfo{year}{2022}), \bibinfo{pages}{647--663}.
\newblock
\href{https://doi.org/10.1093/biomet/asab047}{doi:\nolinkurl{10.1093/biomet/asab047}}


\bibitem[Howard et~al\mbox{.}(2021)]%
        {howard2021confidence}
\bibfield{author}{\bibinfo{person}{Steven~R. Howard}, \bibinfo{person}{Aaditya
  Ramdas}, \bibinfo{person}{Jon McAuliffe}, {and} \bibinfo{person}{Jasjeet
  Sekhon}.} \bibinfo{year}{2021}\natexlab{}.
\newblock \showarticletitle{Time-Uniform, Nonparametric, Nonasymptotic
  Confidence Sequences}.
\newblock \bibinfo{journal}{\emph{The Annals of Statistics}}
  \bibinfo{volume}{49}, \bibinfo{number}{2} (\bibinfo{year}{2021}),
  \bibinfo{pages}{1055--1080}.
\newblock
\href{https://doi.org/10.1214/20-AOS1991}{doi:\nolinkurl{10.1214/20-AOS1991}}


\bibitem[Javanmard and Montanari(2018)]%
        {javanmard2018onlinefdr}
\bibfield{author}{\bibinfo{person}{Adel Javanmard} {and}
  \bibinfo{person}{Andrea Montanari}.} \bibinfo{year}{2018}\natexlab{}.
\newblock \showarticletitle{Online Rules for Control of False Discovery Rate
  and False Discovery Exceedance}.
\newblock \bibinfo{journal}{\emph{The Annals of Statistics}}
  \bibinfo{volume}{46}, \bibinfo{number}{2} (\bibinfo{year}{2018}),
  \bibinfo{pages}{526--554}.
\newblock
\href{https://doi.org/10.1214/17-AOS1559}{doi:\nolinkurl{10.1214/17-AOS1559}}


\bibitem[Lakonishok and Smidt(1988)]%
        {lakonishok1988seasonal}
\bibfield{author}{\bibinfo{person}{Josef Lakonishok} {and}
  \bibinfo{person}{Seymour Smidt}.} \bibinfo{year}{1988}\natexlab{}.
\newblock \showarticletitle{Are Seasonal Anomalies Real? A Ninety-Year
  Perspective}.
\newblock \bibinfo{journal}{\emph{The Review of Financial Studies}}
  \bibinfo{volume}{1}, \bibinfo{number}{4} (\bibinfo{year}{1988}),
  \bibinfo{pages}{403--425}.
\newblock
\href{https://doi.org/10.1093/rfs/1.4.403}{doi:\nolinkurl{10.1093/rfs/1.4.403}}


\bibitem[McLean and Pontiff(2016)]%
        {mclean2016predictability}
\bibfield{author}{\bibinfo{person}{R.~David McLean} {and}
  \bibinfo{person}{Jeffrey Pontiff}.} \bibinfo{year}{2016}\natexlab{}.
\newblock \showarticletitle{Does Academic Research Destroy Stock Return
  Predictability?}
\newblock \bibinfo{journal}{\emph{The Journal of Finance}}
  \bibinfo{volume}{71}, \bibinfo{number}{1} (\bibinfo{year}{2016}),
  \bibinfo{pages}{5--32}.
\newblock
\href{https://doi.org/10.1111/jofi.12365}{doi:\nolinkurl{10.1111/jofi.12365}}


\bibitem[Ramdas et~al\mbox{.}(2023)]%
        {ramdas2023savi}
\bibfield{author}{\bibinfo{person}{Aaditya Ramdas}, \bibinfo{person}{Peter
  Gr{\"u}nwald}, \bibinfo{person}{Vladimir Vovk}, {and} \bibinfo{person}{Glenn
  Shafer}.} \bibinfo{year}{2023}\natexlab{}.
\newblock \showarticletitle{Game-Theoretic Statistics and Safe Anytime-Valid
  Inference}.
\newblock \bibinfo{journal}{\emph{Statist. Sci.}} \bibinfo{volume}{38},
  \bibinfo{number}{4} (\bibinfo{year}{2023}), \bibinfo{pages}{576--601}.
\newblock
\href{https://doi.org/10.1214/23-STS894}{doi:\nolinkurl{10.1214/23-STS894}}


\bibitem[Ramdas et~al\mbox{.}(2018)]%
        {ramdas2018saffron}
\bibfield{author}{\bibinfo{person}{Aaditya Ramdas}, \bibinfo{person}{Tijana
  Zrnic}, \bibinfo{person}{Martin~J. Wainwright}, {and}
  \bibinfo{person}{Michael~I. Jordan}.} \bibinfo{year}{2018}\natexlab{}.
\newblock \showarticletitle{{SAFFRON}: An Adaptive Algorithm for Online Control
  of the False Discovery Rate}. In \bibinfo{booktitle}{\emph{Proceedings of the
  35th International Conference on Machine Learning}}
  \emph{(\bibinfo{series}{Proceedings of Machine Learning Research},
  Vol.~\bibinfo{volume}{80})}. \bibinfo{publisher}{PMLR},
  \bibinfo{pages}{4286--4294}.
\newblock
\urldef\tempurl%
\url{https://proceedings.mlr.press/v80/ramdas18a.html}
\showURL{%
\tempurl}


\bibitem[Romano and Wolf(2005)]%
        {romano2005stepwise}
\bibfield{author}{\bibinfo{person}{Joseph~P. Romano} {and}
  \bibinfo{person}{Michael Wolf}.} \bibinfo{year}{2005}\natexlab{}.
\newblock \showarticletitle{Stepwise Multiple Testing as Formalized Data
  Snooping}.
\newblock \bibinfo{journal}{\emph{Econometrica}} \bibinfo{volume}{73},
  \bibinfo{number}{4} (\bibinfo{year}{2005}), \bibinfo{pages}{1237--1282}.
\newblock
\href{https://doi.org/10.1111/j.1468-0262.2005.00615.x}{doi:\nolinkurl{10.1111/j.1468-0262.2005.00615.x}}


\bibitem[Shafer et~al\mbox{.}(2011)]%
        {shafer2011testmartingales}
\bibfield{author}{\bibinfo{person}{Glenn Shafer}, \bibinfo{person}{Alexander
  Shen}, \bibinfo{person}{Nikolai Vereshchagin}, {and}
  \bibinfo{person}{Vladimir Vovk}.} \bibinfo{year}{2011}\natexlab{}.
\newblock \showarticletitle{Test Martingales, Bayes Factors and p-Values}.
\newblock \bibinfo{journal}{\emph{Statist. Sci.}} \bibinfo{volume}{26},
  \bibinfo{number}{1} (\bibinfo{year}{2011}), \bibinfo{pages}{84--101}.
\newblock
\href{https://doi.org/10.1214/10-STS347}{doi:\nolinkurl{10.1214/10-STS347}}


\bibitem[Sullivan et~al\mbox{.}(1999)]%
        {sullivan1999technical}
\bibfield{author}{\bibinfo{person}{Ryan Sullivan}, \bibinfo{person}{Allan
  Timmermann}, {and} \bibinfo{person}{Halbert White}.}
  \bibinfo{year}{1999}\natexlab{}.
\newblock \showarticletitle{Data-Snooping, Technical Trading Rule Performance,
  and the Bootstrap}.
\newblock \bibinfo{journal}{\emph{The Journal of Finance}}
  \bibinfo{volume}{54}, \bibinfo{number}{5} (\bibinfo{year}{1999}),
  \bibinfo{pages}{1647--1691}.
\newblock
\href{https://doi.org/10.1111/0022-1082.00163}{doi:\nolinkurl{10.1111/0022-1082.00163}}


\bibitem[Sullivan et~al\mbox{.}(2001)]%
        {sullivan2001calendar}
\bibfield{author}{\bibinfo{person}{Ryan Sullivan}, \bibinfo{person}{Allan
  Timmermann}, {and} \bibinfo{person}{Halbert White}.}
  \bibinfo{year}{2001}\natexlab{}.
\newblock \showarticletitle{Dangers of Data Mining: The Case of Calendar
  Effects in Stock Returns}.
\newblock \bibinfo{journal}{\emph{Journal of Econometrics}}
  \bibinfo{volume}{105}, \bibinfo{number}{1} (\bibinfo{year}{2001}),
  \bibinfo{pages}{249--286}.
\newblock
\href{https://doi.org/10.1016/S0304-4076(01)00077-X}{doi:\nolinkurl{10.1016/S0304-4076(01)00077-X}}


\bibitem[Ville(1939)]%
        {ville1939}
\bibfield{author}{\bibinfo{person}{Jean Ville}.}
  \bibinfo{year}{1939}\natexlab{}.
\newblock \bibinfo{booktitle}{\emph{{\'E}tude critique de la notion de
  collectif}}.
\newblock \bibinfo{publisher}{Gauthier-Villars}, \bibinfo{address}{Paris}.
\newblock


\bibitem[Vovk and Wang(2021)]%
        {vovk2021evalues}
\bibfield{author}{\bibinfo{person}{Vladimir Vovk} {and} \bibinfo{person}{Ruodu
  Wang}.} \bibinfo{year}{2021}\natexlab{}.
\newblock \showarticletitle{E-values: Calibration, Combination, and
  Applications}.
\newblock \bibinfo{journal}{\emph{The Annals of Statistics}}
  \bibinfo{volume}{49}, \bibinfo{number}{3} (\bibinfo{year}{2021}),
  \bibinfo{pages}{1736--1754}.
\newblock
\href{https://doi.org/10.1214/20-AOS2020}{doi:\nolinkurl{10.1214/20-AOS2020}}


\bibitem[Wang et~al\mbox{.}(2025)]%
        {wang2025ebacktesting}
\bibfield{author}{\bibinfo{person}{Qiuqi Wang}, \bibinfo{person}{Ruodu Wang},
  {and} \bibinfo{person}{Johanna Ziegel}.} \bibinfo{year}{2025}\natexlab{}.
\newblock \showarticletitle{E-backtesting}.
\newblock \bibinfo{journal}{\emph{Management Science}} \bibinfo{volume}{72},
  \bibinfo{number}{6} (\bibinfo{year}{2025}), \bibinfo{pages}{4952--4973}.
\newblock
\shownote{Published online September 23, 2025; issue dated June 2026}.
\newblock
\href{https://doi.org/10.1287/mnsc.2023.01659}{doi:\nolinkurl{10.1287/mnsc.2023.01659}}


\bibitem[Wang and Ramdas(2022)]%
        {wang2022ebh}
\bibfield{author}{\bibinfo{person}{Ruodu Wang} {and} \bibinfo{person}{Aaditya
  Ramdas}.} \bibinfo{year}{2022}\natexlab{}.
\newblock \showarticletitle{False Discovery Rate Control with E-values}.
\newblock \bibinfo{journal}{\emph{Journal of the Royal Statistical Society
  Series B: Statistical Methodology}} \bibinfo{volume}{84}, \bibinfo{number}{3}
  (\bibinfo{year}{2022}), \bibinfo{pages}{822--852}.
\newblock
\href{https://doi.org/10.1111/rssb.12489}{doi:\nolinkurl{10.1111/rssb.12489}}


\bibitem[Waudby-Smith and Ramdas(2024)]%
        {waudbysmith2024betting}
\bibfield{author}{\bibinfo{person}{Ian Waudby-Smith} {and}
  \bibinfo{person}{Aaditya Ramdas}.} \bibinfo{year}{2024}\natexlab{}.
\newblock \showarticletitle{Estimating Means of Bounded Random Variables by
  Betting}.
\newblock \bibinfo{journal}{\emph{Journal of the Royal Statistical Society
  Series B: Statistical Methodology}} \bibinfo{volume}{86}, \bibinfo{number}{1}
  (\bibinfo{year}{2024}), \bibinfo{pages}{1--27}.
\newblock
\href{https://doi.org/10.1093/jrsssb/qkad009}{doi:\nolinkurl{10.1093/jrsssb/qkad009}}


\bibitem[White(2000)]%
        {white2000reality}
\bibfield{author}{\bibinfo{person}{Halbert White}.}
  \bibinfo{year}{2000}\natexlab{}.
\newblock \showarticletitle{A Reality Check for Data Snooping}.
\newblock \bibinfo{journal}{\emph{Econometrica}} \bibinfo{volume}{68},
  \bibinfo{number}{5} (\bibinfo{year}{2000}), \bibinfo{pages}{1097--1126}.
\newblock
\href{https://doi.org/10.1111/1468-0262.00152}{doi:\nolinkurl{10.1111/1468-0262.00152}}


\bibitem[Yuan et~al\mbox{.}(2026)]%
        {yuan2026metatrader}
\bibfield{author}{\bibinfo{person}{Haochen Yuan}, \bibinfo{person}{Minting
  Pan}, \bibinfo{person}{Yunbo Wang}, \bibinfo{person}{Siyu Gao}, {and}
  \bibinfo{person}{Xiaokang Yang}.} \bibinfo{year}{2026}\natexlab{}.
\newblock \showarticletitle{{MetaTrader}: Learning to Generalize {RL} Trading
  Policies Beyond Offline Data}. In \bibinfo{booktitle}{\emph{Proceedings of
  the AAAI Conference on Artificial Intelligence}}, Vol.~\bibinfo{volume}{40}.
  \bibinfo{pages}{28023--28031}.
\newblock
\href{https://doi.org/10.1609/aaai.v40i33.40027}{doi:\nolinkurl{10.1609/aaai.v40i33.40027}}


\bibitem[Zrnic et~al\mbox{.}(2021)]%
        {zrnic2021asynchronous}
\bibfield{author}{\bibinfo{person}{Tijana Zrnic}, \bibinfo{person}{Aaditya
  Ramdas}, {and} \bibinfo{person}{Michael~I. Jordan}.}
  \bibinfo{year}{2021}\natexlab{}.
\newblock \showarticletitle{Asynchronous Online Testing of Multiple
  Hypotheses}.
\newblock \bibinfo{journal}{\emph{Journal of Machine Learning Research}}
  \bibinfo{volume}{22}, \bibinfo{number}{33} (\bibinfo{year}{2021}),
  \bibinfo{pages}{1--39}.
\newblock
\urldef\tempurl%
\url{https://jmlr.org/papers/v22/19-910.html}
\showURL{%
\tempurl}


\end{thebibliography}

\end{document}